\documentclass[conference]{IEEEtran}
\usepackage{times}

\usepackage[numbers]{natbib}
\usepackage{multicol}
\usepackage[bookmarks=true]{hyperref}
\usepackage{graphics} 
\usepackage{epsfig} 
\usepackage{amsmath} 
\usepackage{amssymb}  
\usepackage{relsize}
\usepackage{hyperref}

\begin{document}

\title{Interpreting and Predicting Tactile Signals via a Physics-Based and Data-Driven Framework}

\author{\authorblockN{Yashraj S. Narang\authorrefmark{1},
Karl Van Wyk\authorrefmark{1},
Arsalan Mousavian\authorrefmark{1}, and 
Dieter Fox\authorrefmark{1}\authorrefmark{2}}
\authorblockA{\authorrefmark{1}NVIDIA Corporation, Seattle, USA\\
Email: ynarang@nvidia.com}
\authorblockA{\authorrefmark{2}Paul G. Allen School of Computer Science \& Engineering\\
University of Washington,
Seattle, USA}}

\maketitle

\begin{abstract}
High-density afferents in the human hand have long been regarded as essential for human grasping and manipulation abilities. In contrast, robotic tactile sensors are typically used to provide low-density contact data, such as center-of-pressure and resultant force. Although useful, this data does not exploit the rich information content that some tactile sensors (e.g., the SynTouch BioTac) naturally provide. This research extends robotic tactile sensing beyond reduced-order models through 1) the automated creation of a precise tactile dataset for the BioTac over diverse physical interactions, 2) a 3D finite element (FE) model of the BioTac, which complements the experimental dataset with high-resolution, distributed contact data, and 3) neural-network-based mappings from raw BioTac signals to low-dimensional experimental data, and more importantly, high-density FE deformation fields. These data streams can provide a far greater quantity of interpretable information for grasping and manipulation algorithms than previously accessible.
\end{abstract}

\IEEEpeerreviewmaketitle

\section{Introduction}

There are three major sensing modalities in robotic grasping and manipulation: proprioception, vision, and tactile sensing. Among these modalities, tactile sensing provides the most direct information about the physical properties of the object during interaction, including mass, stiffness, friction, and surface texture, as well as information about contact locations and forces. In addition, tactile data is essential for grasping and manipulation in the presence of visual occlusion, including robot self-occlusion, deep object concavities, and clutter \cite{Mason2018AnnRev, Billard2019Science}. A large number of compelling tactile sensors have been developed for robotic applications. These sensors include the GelSight\cite{Yuan2017Sensors}, GelSlim~\cite{Donlon2018IROS}, TacTip\cite{Ward2018SoRo}, and commercial offerings from SynTouch and Pressure Profile Systems. For recent reviews, see \cite{Dahiya2010TRO, Yousef2011SensAct, KapassovRAS2015, Chen2018Sensors, Akihiko2019AdvRob}.

In robotics, tactile sensing has become especially useful for tasks such as object identification and slip detection\cite{Luo2017Mechatronics, Chen2018Sensors}. For general grasping and manipulation, key challenges remain, including 1) How does one accurately estimate low-dimensional \textit{tactile features}, such as center of pressure and force vector, from raw tactile signals? These features facilitate classical grasping and manipulation methods, such as assessing grasp stability and planning hand-finger trajectories for rigid objects \cite{Murray1994Book, Mason2001Book}; and 2) How does one estimate high-resolution \textit{tactile fields} (e.g., local deformations of the surface of the tactile sensor) from raw tactile signals? These fields provide high-density information useful for interacting with small, geometrically-irregular, fragile, and/or compliant objects \cite{Bicchi2000ICRA, Li2001ICRA}. 

\begin{figure}[t]
  \centering
  \framebox{\parbox{2.8in}{\includegraphics[scale=0.225]{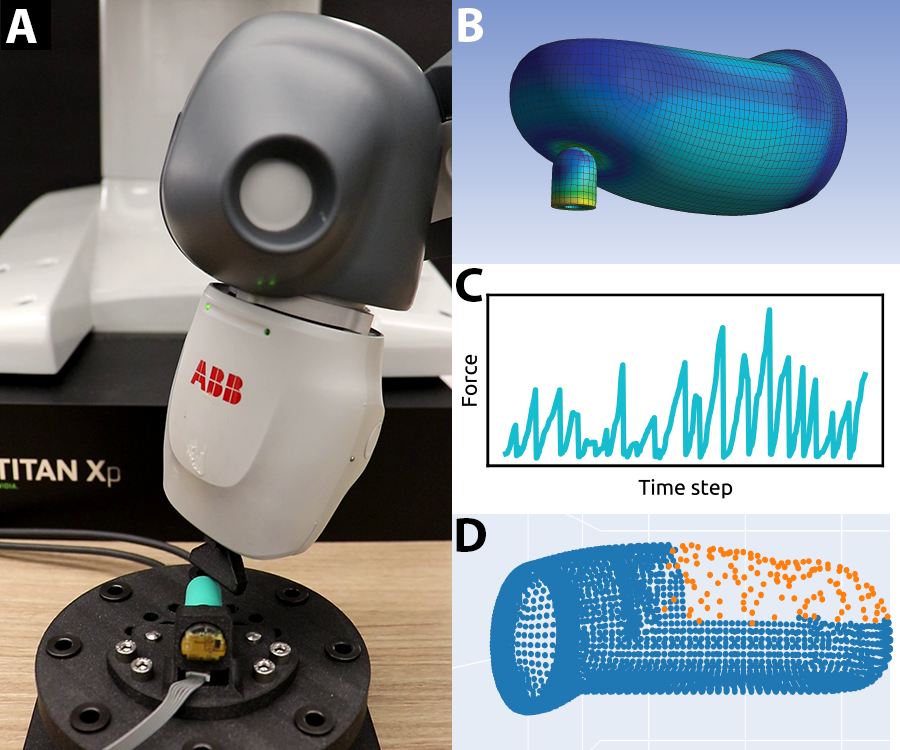}}}
  \caption{This research presents the following for the SynTouch BioTac: A) a large experimental dataset, B) the first 3D finite element model, C) regression to tactile features, and D) the first regression to dense tactile fields.}
  \label{intro}
\end{figure}

This paper focuses on addressing these questions for a canonical tactile sensor, the SynTouch BioTac (Fig. \ref{intro}). The BioTac is a fingertip-shaped sensor with a fluid-coupled electrode array that measures impedances complexly related to surface deformations and distributed forces. The BioTac was selected due to its high spatial resolution and sensitivity, low stiffness and hysteresis, and widespread use in research \cite{Wettels2008AdvRob, Ruppel2019IAS}. The primary contributions\footnote{Datasets, CAD files for the testbed, FE model files, and a brief summary video are available at \textit{\url{https://sites.google.com/nvidia.com/tactiledata}.}} of our work are

\begin{enumerate}
    \item A novel experimental dataset containing raw electrode values, contact locations, and force vectors. The dataset was collected from nine robot-controlled indenters interacting with three BioTac sensors, with over $300$ unique indentation trajectories, $700$ total trajectories, and $30k$ data points after processing. Contact location was precisely measured through careful design and calibration of a novel testbed. Approximately $70\%$ of the trajectories were designed to induce shear forces, a critical mechanical phenomenon in grasping and manipulation.
    
    \item The first 3D finite-element (FE) model of the BioTac. Although previous studies have hypothesized that ``the deflection of the rubber skin [of the BioTac] is almost impossible to model'' \cite{Ruppel2019IAS}, the FE model captures this behavior and its underlying mechanical phenomena. This model was validated against experimental force vectors and generalized over the wide range of indenters and contact locations; thus, the model can be used to predict the mechanical behavior of the sensor in diverse conditions. FE predictions of surface deformations are also included in the previously-described dataset. 
    
    \item Neural-network mappings between A) electrode signals and tactile features, and B) electrode signals and tactile fields. Mapping A can be used to implement classical grasping and manipulation methods using raw tactile data. Mapping B is a first-of-its-kind, high-density version of Mapping A, which can inform control in far greater detail, as well as motivate the design of novel algorithms to leverage such high-resolution information.
\end{enumerate} 

\section{Related Works}

This section reviews previous efforts to A) estimate contact location and resultant force vector from electrode signals on the BioTac, and B) model or estimate local displacement fields on the sensor or object from electrode signals.

\subsection{Tactile Feature Estimation}

Wettels and Loeb estimated contact location and force vector on a BioTac prototype\cite{Wettels2011Robio}. The device was manually indented using four objects of varying curvature. Contact location was estimated using electrode signals in a weighted average, achieving an RMS error of $2.4$-$2.9$~$mm$, and force vector was estimated using a multilayer perceptron (MLP), achieving an error of $18$-$40\%$. Wettels and Loeb later found that Gaussian mixture models were less accurate \cite{Wettels2014Springer}. Su, et al. (2012) estimated force vector on the BioTac. The device was manually manipulated, and force vector was estimated using electrode signals and normal vectors in a weighted sum, achieving an RMS error of $0.42$-$0.48$ N (up to $10\%$) along each spatial axis\cite{Su2012FrontInNeurorob}. Critically, the weighted sum assumed that a given electrode provides information about force only along its normal vector. Lin, et al. estimated contact location and force vector \cite{Lin2013TechRep}. The BioTac was manually indented using a flat cylindrical indenter. Similar to \cite{Wettels2011Robio}, contact location was estimated using a weighted average and then projected to an approximate geometric model of the BioTac surface. Force vector was estimated using the method from \cite{Su2012FrontInNeurorob}. Su, et al. (2015) estimated force vector. The BioTac was manually indented by a human finger. Force vector was estimated using four different methods: the analytical method of \cite{Su2012FrontInNeurorob}, locally-weighted projection regression, the MLP of \cite{Wettels2011Robio}, and a five-layer neural network \cite{Su2015Humanoids}. The five-layer network was most accurate, achieving an RMS error of $0.43$-$0.85$ N along each spatial axis. Most recently, Sundaralingam, et al. estimated force vector \cite{Sundaralingam2019ICRA}. The BioTac was 1) manually indented with a large flat object, 2) attached to a robotic hand to contact a sphere, and 3) attached to the same hand to push a box. Critically, forces were primarily applied in the normal direction, and contact location was not measured. Force vector was estimated using a 3D convolutional neural network (CNN), achieving median force magnitude errors of $0.32$-$0.51$ N (35-40\%) and direction errors of $0.07$-$0.39$ rad.

In summary, the highest-quality existing datasets for the BioTac involve manual indentation of a single sensor, imprecise measurement of contact location, and/or forces applied in the normal direction. Using these datasets, the state-of-the-art in tactile feature estimation is 1) RMS contact location error of $2$-$3$~$mm$, 2) median force magnitude error of $0.3$-$0.5$ N, and 3) median force direction error of $0.07$ rad.

\subsection{Tactile Field Estimation}

No prior studies were found that modeled or estimated local displacement fields on the surface of the BioTac or a contacting object. The only related effort was by Wettels, et al., which modeled the indentation of internal fluidic channels of the BioTac\cite{Wettels2008BioRob}. A 2D plane-stress FE simulation was generated, and two channel geometries were qualitatively evaluated.

\section{Methods}

\subsection{Experimental Testbed Design}

\begin{figure}[thpb]
  \centering
  \framebox{\parbox{2.8in}{\includegraphics[scale=.8]{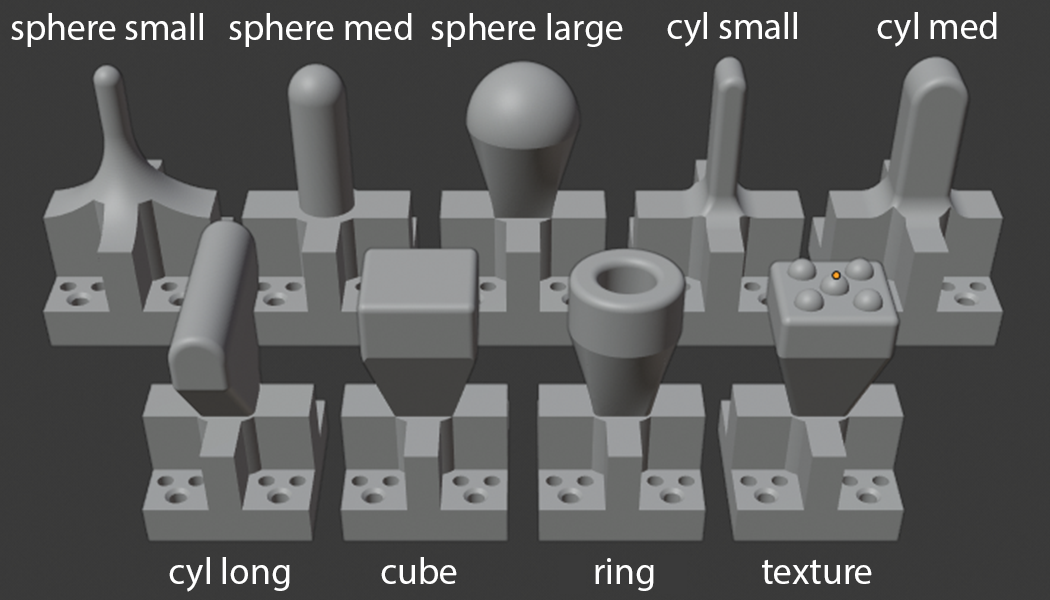}}}
  \caption{Nine indenters capturing primitive geometry of everyday objects. Examples include \textit{cylinder large}, modeling table edges; \textit{texture}, modeling surfaces with scattered debris; and \textit{ring}, modeling bottle openings. Critical dimensions were approximately half, equal, or double the radius of the BioTac.}
  \label{indenters}
\end{figure}

To automatically collect high-quality ground-truth contact locations and 3D force vectors from the BioTac, a custom testbed was designed and carefully calibrated. The testbed consisted of 1) an ABB YuMi bimanual robot, 2) 3D-printed indenters attached to the distal links of the robot, and 3) a mount that rigidly coupled the BioTac to a Weiss Robotics KMS40 6-axis force/torque (F/T) sensor. The YuMi was selected due to its high positional repeatability ($0.02$~$mm$) and large dexterous workspace. The indenters were designed to capture primitive geometry of everyday household objects (Fig. \ref{indenters}). They were 3D-printed using a Prusa i3 MK3 printer at fine resolution ($0.05$~$mm$) and attached to the distal links of the YuMi with precision alignment pins.

\begin{figure}[thpb]
  \centering
  \framebox{\parbox{3.25in}{\includegraphics[scale=0.9]{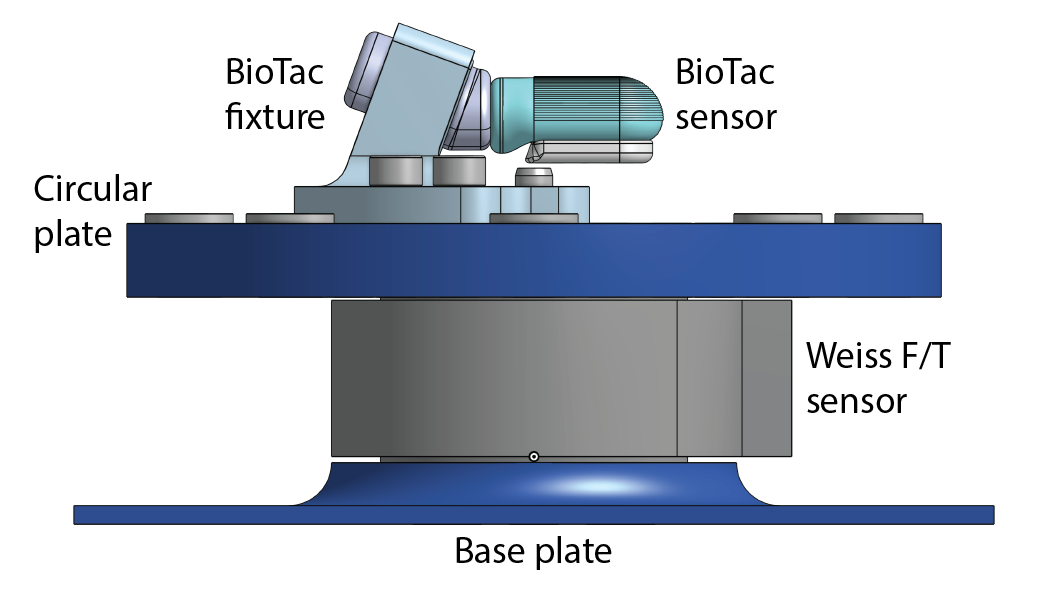}}}
  \caption{Schematic of mount used in experimental testbed (side view)}
  \label{assembly}
\end{figure}

The mount itself consisted of 1) a 3D-printed fixture into which the BioTac was inserted, 2) a circular 3D-printed plate that coupled the BioTac fixture to the F/T sensor, 3) the F/T sensor itself, and 4) a base plate that coupled the F/T sensor to a benchtop (Fig. \ref{assembly}). The BioTac achieved an interference fit within its fixture and was secured at a predefined depth using a set screw. The fixture, circular plate, F/T sensor, and base plate were aligned using precision dowel pins.

\subsection{Mechanical Registration Procedure}

\begin{figure}[thpb]
  \centering
  \framebox{\parbox{2.93in}{\includegraphics[scale=0.2]{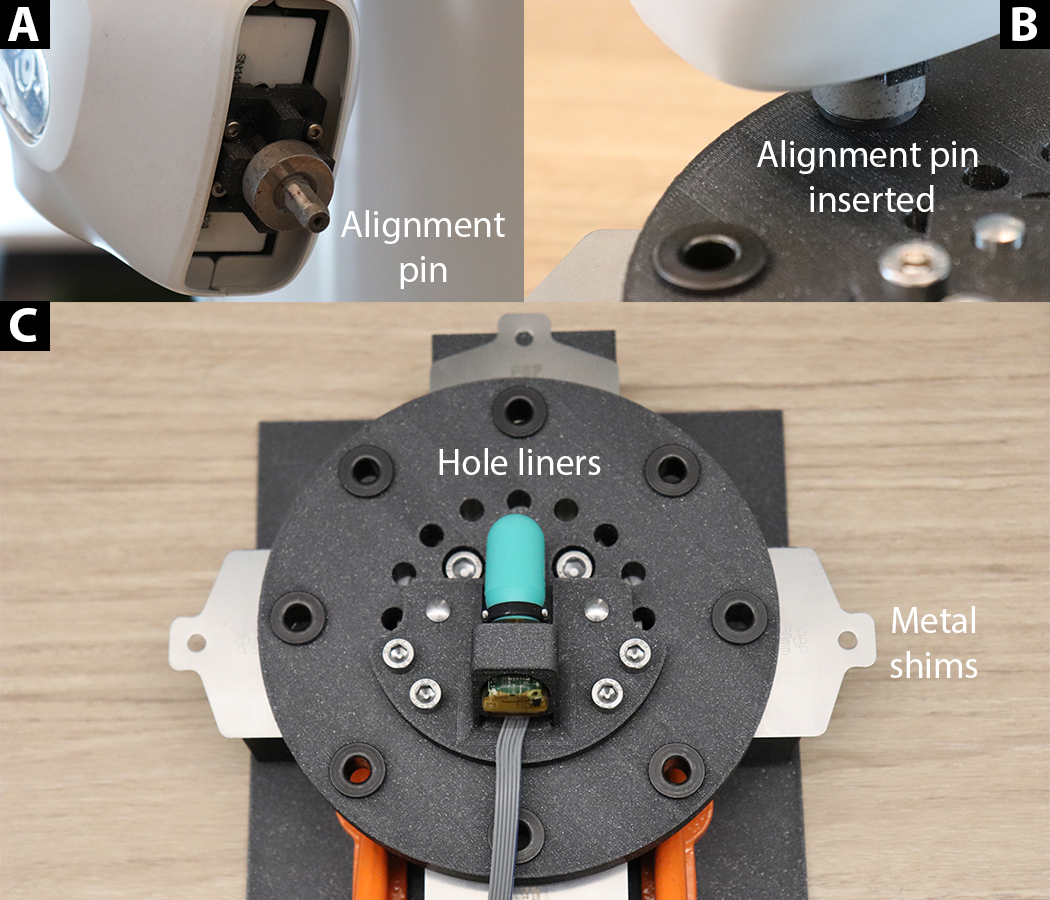}}}
  \caption{Mechanical calibration apparatus. A) Precision alignment pin attached to end effector. B) Alignment pin inserted into precision metal hole liner during registration. C) Overhead view of mount, illustrating liners and shims.}
  \label{setup}
\end{figure}

To spatially register the 6D pose of the robot end-effectors with the BioTac, a mechanical calibration apparatus was designed. The apparatus consisted of a diamond-head shoulder-style metal alignment pin attached to the distal link of the YuMi robot (Fig. \ref{setup}A), and metal hole liners press-fit into holes along the circumference of the circular plate (Fig. \ref{setup}C). The worst-case clearance between the alignment pin and hole liners was approximately $0.3$~$mm$. Metal shims were temporarily inserted between the circular plate and the F/T sensor during calibration to prevent pitching of the plate.

During registration, lead-through mode on the YuMi robot was enabled, and the alignment pin was guided into a metal hole liner until shoulder contact (Fig. \ref{setup}B). Upon contact, end-effector coordinates were recorded in a robot-fixed coordinate frame $R$ (using joint angles and a forward kinematic model) and a BioTac-fixed coordinate frame $B$ (using a CAD model of the BioTac and experimental mount). This step was repeated for two additional hole liners. Following the algorithm outlined in \cite{Marvel2016ISAM}, these three non-collinear points were used to construct a new intermediate coordinate system $I$, which could then be expressed in the robot and BioTac coordinate frames through homogeneous transformation matrices ${}^R_I H$ and ${}^B_I H$. The transformation matrix ${}^B_R H$ from the robot to BioTac coordinate frames was then simply ${}^B_I H ({}^R_I H)^{-1}$.

\begin{figure}[thpb]
  \centering
  \framebox{\parbox{3.25in}{\includegraphics[scale=0.224]{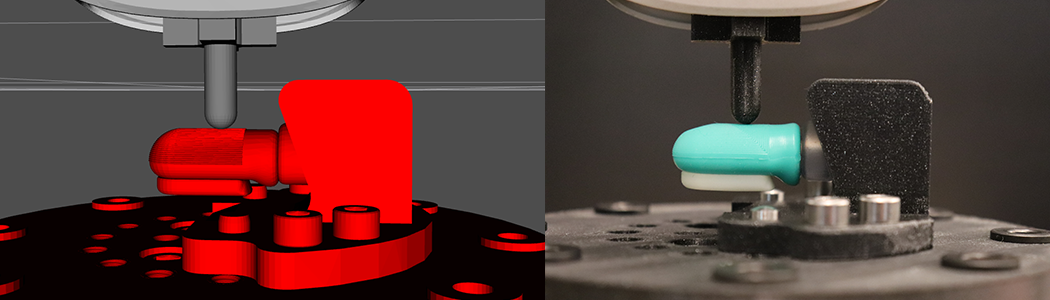}}}
  \caption{Example of mechanical registration performance. Left: Predicted poses of \textit{sphere medium} indenter and BioTac, with indenter just initiating contact with BioTac. Right: Observed poses.}
  \label{registration}
\end{figure}

To ensure robustness to measurement error, the previous procedure was repeated for all three-hole combinations of four different hole liners on the circular plate. The transformation matrices for all combinations were then averaged to produce a final transformation matrix, as proposed and validated in \cite{VanWyk2018TASE}. Registration error between the robot and BioTac was observed to be between $0.5$-$1.5$~$mm$ (Fig. \ref{registration}) over the full range of robot joint configurations and end-effector positions traversed during testing. The full mechanical registration process was applied to the right and left arms of the YuMi robot independently.

\subsection{Experimental Testing Procedure}

For each of the nine indenters (Fig. \ref{indenters}), ten unique points were randomly sampled from the ventral surface of the BioTac. Points to the right of the long axis of the finger were assigned to the right arm of the YuMi robot, and vice versa for the left. For each point, an indentation trajectory was generated normal to the surface. In addition, four angled trajectories were generated at $30$~$deg$ from the normal to purposely induce shear. Each trajectory indented the sensor by $3$~$mm$ (for normal indentations) or $1.5$~$mm$ (for angled indentations) past initial contact. In addition, each trajectory was divided into $0.1$~$mm$ displacement increments, with a 5-second pause after each increment to allow the BioTac fluid to settle. Joint commands were generated using Riemannian Motion Policies due to their efficient avoidance of self-collision and specified obstacles\cite{Ratliff2018ArXiv}, and trajectories were simulated using ROS \textit{rviz}. Trajectories were eliminated that could not be achieved with high accuracy (i.e., $\leq$ $0.01$~$mm$, according to a forward-kinematic model) or resulted in unwanted collisions.

All trajectories were then executed on the real-world experimental setup for three different BioTac sensors. However, one sensor (\textit{BioTac 1}) had data exclusively from the \textit{sphere small}, \textit{sphere medium}, and \textit{sphere large} indenters due to damage during testing. Joint angles from the active YuMi arm ($\mathbb{R}^{7}$), force data from the F/T sensor ($\mathbb{R}^{3}$), and electrode values from the BioTac ($\mathbb{R}^{19}$) were continuously acquired at $\geq$~$100$~$Hz$.

\subsection{Experimental Data Processing}

All data were time-aligned in post-processing. Robot joint angles were converted to indenter tip positions using a forward-kinematic model. To mitigate drift, BioTac electrode values and force/torque values for each indentation were tared against their corresponding values at the first time step of the trajectory (i.e., before contact with the indenter). Due to noise on the z-axis readings of the force/torque sensor, a first-order Butterworth low-pass filter with a cutoff frequency of $5$~$Hz$ was applied forward and backward to the force data. To reduce the dataset size ($>$~$1.5e6$ samples) and mitigate redundancy, the datasets were subsampled at the final time step of each $0.1$~$mm$ displacement increment within each trajectory.

\subsection{Finite Element Modeling}

For local tactile fields, such as deformation fields on the surface of the BioTac, real-world ground-truth data is prohibitively difficult to acquire. To generate near-ground-truth data for such quantities, an FE model was developed using ANSYS Mechanical v19.1. The FE method is a variational numerical technique that divides complex geometric domains into simple subregions and solves the weak form of the governing PDEs over each region. With sufficiently high mesh density and small time steps, predictions for the deformation of solids can be extremely accurate\cite{Reddy2019Book}.

\begin{figure}[thpb]
  \centering
  \framebox{\parbox{3.25in}{\includegraphics[scale=0.224]{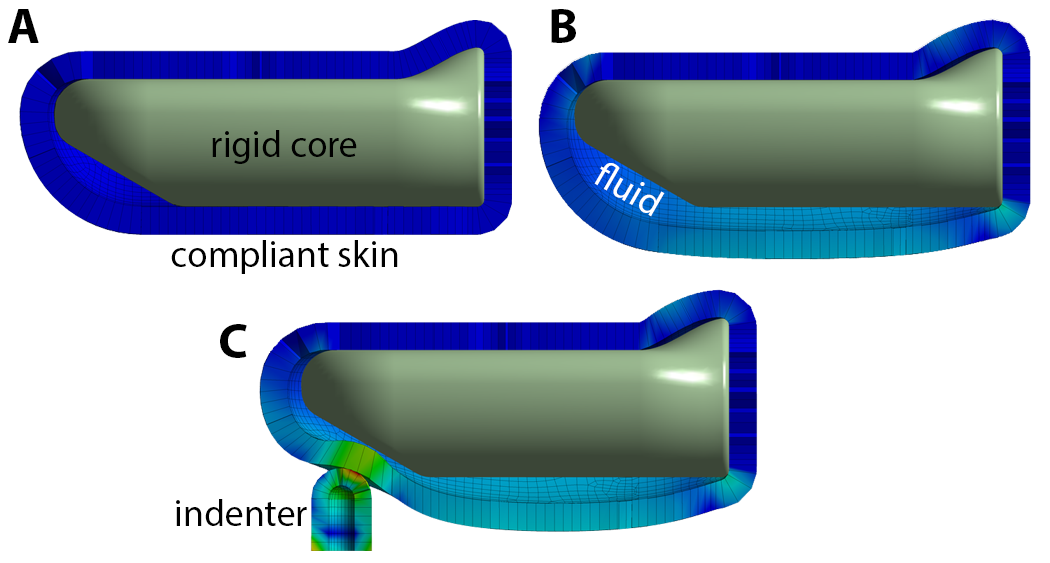}}}
  \caption{Cross-sectional views of BioTac sensor during example FE simulation. A) Initial state. B) After pressurization. C) During indentation. Colors illustrate von Mises equivalent stress.}
  \label{ansys}
\end{figure}

Geometrically, the finite element model consisted of 1) the BioTac skin, 2) the BioTac core, 3) the fluid between the skin and core, and 4) the indenter (Fig. \ref{ansys}). The initial CAD models of the skin and core were acquired from the manufacturer; however, the asperities on the skin (e.g., fingerprints) were removed to facilitate meshing and expedite convergence. Four-node shell elements (SHELL181) were selected for the skin due to their high efficiency and accuracy for modeling large, nonlinear deformation of membranes\cite{Lee2018Book}. The core was defined as a rigid body, and hydrostatic fluid elements (HSFLD242) were used for the fluid layer to efficiently enforce incompressibility (i.e., conservation of volume).

Constitutive laws (i.e., material parameters) for the components of the model were then defined. The skin of the BioTac is made of silicone rubber; such materials can be accurately modeled with a hyperelastic material model, in which the stress-strain relationships are derived from a Helmholtz free energy density function\cite{Treloar1974Book}. Due to its mathematical simplicity and accuracy over low stretch, the incompressible Neo-Hookean energy density function was chosen, for which $\mu_{NH}$ is the unknown stiffness parameter. In addition, although the BioTac may be approximated as isothermal, an arbitrary volumetric thermal expansion coefficient $\beta_T = 0.01$ $C^{-1}$ was defined for the fluid as a simple means to later adjust its volume.

From physical considerations, three spatial boundary conditions were defined: 1) grounding of the rigid core, 2) anchoring together the proximal (i.e., right, in Fig. \ref{ansys}) surfaces of the skin and core, 3) anchoring together the dorsal (i.e., top, in Fig. \ref{ansys}) surfaces of the skin and core. In addition, two contact boundary conditions were defined: frictionless contact between the skin and core (due to lubrication provided by the fluid), and frictional contact between the indenter and the skin. The normal-Lagrange contact formulation was selected to minimize interpenetration. A mesh refinement study determined that a maximum mesh dimension less than $0.5$~$mm$ produced negligible increases in solution accuracy.

Two load steps were applied: 1) a pressurization step, in which the cavity between the skin and core was inflated with fluid (Fig. \ref{ansys}B), and 2) an indentation step, in which displacement increments were applied to the indenter (Fig. \ref{ansys}C). Step 1 was necessary because the manufacturer only provided CAD models of the unpressurized sensor (i.e., no fluid); in simulation, the temperature of the fluid was increased until the total sensor thickness matched the manufacturer's specification of $15.1$~$mm$\cite{SynTouch2018Manual}. For step 2, the indentation trajectories exactly matched the target trajectories in the experiments.

An asymmetric Newton-Raphson solver was selected to facilitate convergence for frictional contact, and the simulation was executed. Two primary quantities were extracted: \textit{nodal coordinates} at the end of Step 1, which served as reference coordinates, and \textit{nodal displacements} throughout Step 2, relative to the reference coordinates. The preceding simulation and data extraction procedure was repeated for every indenter and target trajectory examined in the experiments.

\subsection{Finite Element Model Calibration}

As FE models are highly constrained by the laws of physics, only a few free parameters need to be measured or estimated. The current model has four parameters: 1) the thickness $t_{sk}$ of the skin, which was modified to remove surface asperities (and thus, requires tuning to achieve accurate deformation relative to the original CAD model), 2) coefficient $\mu_{NH}$ in the Neo-Hookean energy density function, 3) coefficient of friction $\mu_{fr}$ between the indenter and skin, and 4) the fluid temperature $T_{fl}$ that, given parameters 1 and 2, achieves a desired sensor thickness of $15.1$~$mm$ at the end of the first load step.

\begin{figure}[thpb]
  \centering
  \framebox{\parbox{2.75in}{\includegraphics[scale=0.9]{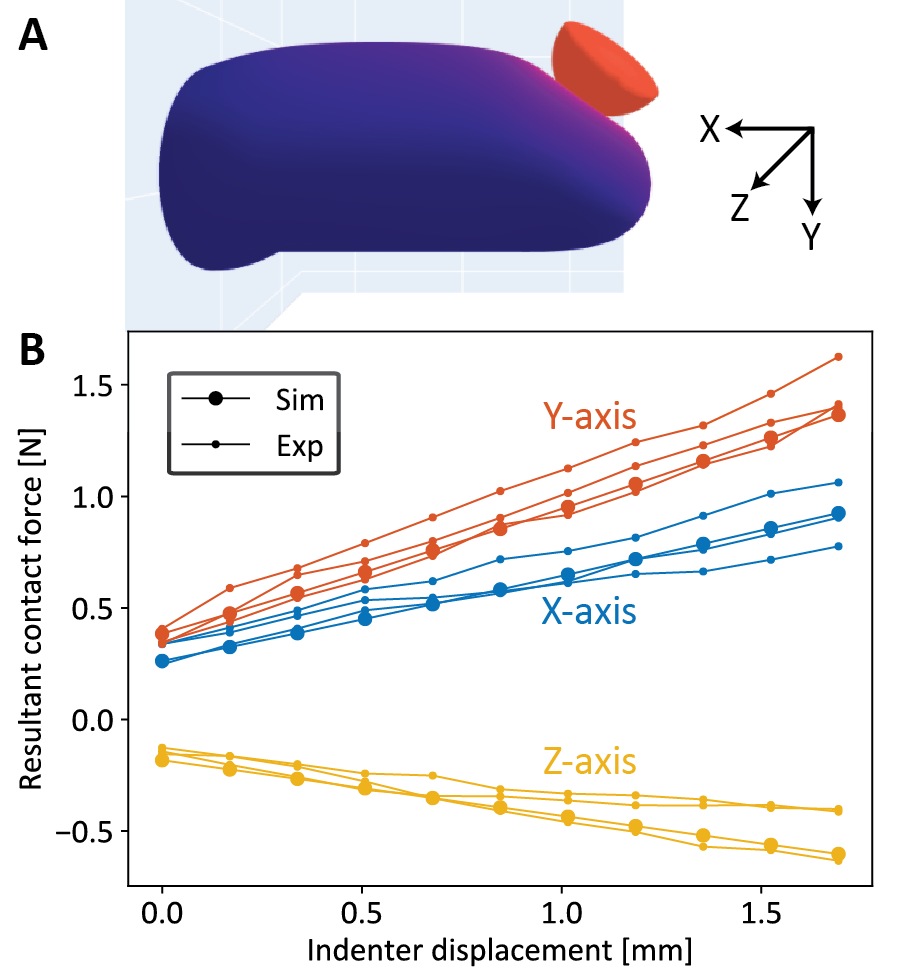}}}
  \caption{FE calibration results. A) Angled trajectory used for calibration of FE parameters. B) Simulated and experimental data for contact force along each axis. Experimental data from three different BioTac sensors is shown.}
  \label{optim_calib}
\end{figure}

To calibrate the values of these parameters, the resultant contact force on the BioTac was compared between FE simulations and corresponding experiments, and the parameters were tuned to minimize MSE over all time steps. Due to the small number of parameters, it was hypothesized that minimizing force error for a \textit{single} indentation trajectory would produce parameters that generalize well across \textit{all} indentations. As follows, a specific angled trajectory of the \textit{sphere medium} indenter was selected due to its moderate contact area, localized deformation, and generation of both normal and shear stresses (Fig. \ref{optim_calib}). A sequential least-squares programming (SLSQP) optimizer was used to minimize the following cost function:
\begin{align}
J_{total} & = w_1 J_{force} + w_2 J_{thick} \label{eq:1} \\
& = w_1 \mathlarger{\sum_{i=1}^3} \sqrt{\frac{1}{N} \sum_{j=1}^N (F_{ij}^{sim}-F_{ij}^{exp})^2} \nonumber \\
& + w_2(t_{sn}^{sim}-0.0151) \nonumber
\end{align}

\noindent where $w_1$ and $w_2$ are hand-selected weights; $J_{force}$ is the sum of the RMS contact force error for each spatial axis; $J_{thick}$ is a cost function ensuring that the total sensor thickness after pressurization matches the manufacturer's specification; $N$ is the total number of samples (i.e., product of number of sensors and time steps per sensor); $F_{ij}^{sim}$ and $F_{ij}^{exp}$ are the simulated and experimental force, respectively, for spatial axis $i$ and sample number $j$; $t_{sn}^{sim}$ is the simulated total sensor thickness; and $0.0151$~$m$ is the manufacturer-specified thickness. Note that $F_{ij}^{sim}$ and $t_{sn}^{sim}$ are functions of simulation parameters $t_s$, $\mu_{NH}$, $\mu_{fr}$, and $T_{fl}$.

Weights $w_1$ and $w_2$ were selected to be $1$ and $1e4$, respectively, such that $J_{force}$ was approximately one order of magnitude higher than $J_{thick}$ throughout optimization. Upper and lower bounds on the simulation parameters were chosen based on physical priors on the materials used in the sensor. The bounds, as well as the optimized values of the parameters, are given in Table~\ref{optim}. FE predictions and experimental values for the net force after optimization are given in Fig. \ref{optim_calib}.

\bgroup
\def\arraystretch{1.5}
\begin{table}[thpb]
\caption{Optimization bounds and optimized values}
\centering{}
\begin{tabular}{|l|l|l|l|l|}
\hline
         & $t_s$ [m] & $\mu_{NH}$ [Pa] & $\mu_{fr}$ & $T_{fl}$ [$^{\circ}$C] \\ \hline
lower      & 1e-3    & 1e5    & 0.1  & 25 \\ \hline
upper      & 2e-3    & 1e6    & 1.0  & 35 \\ \hline
optimal  & 1.57e-3 & 2.80e5 & 0.186 & 29.19 \\ \hline
\end{tabular}
\label{optim}
\end{table}
\egroup

\subsection{Finite Element Model Validation}

To validate the calibrated FE model, the predicted resultant contact force was compared to corresponding experimental values for every single indenter and trajectory. Due to persistent noise from the F/T sensor at low measured forces, only experimental data above $0.5$~$N$ was compared; previous studies have set this threshold higher (e.g., $1.0$~$N$ in \cite{Lin2009Robio}). Furthermore, during contact between the skin and the internal rigid core, simulations often diverged due to the stiffness of the compressed rubber, combined with moderately-large time steps to ensure reasonable simulation time. Only simulation data below the divergence point was compared. Results of the validation are presented in the following section.

\subsection{Tactile Feature Estimation}

Whereas previous neural-network-based efforts to estimate tactile features utilized MLPs\cite{Wettels2011Robio, Su2015Humanoids} and a 3D CNN with sparse occupancy\cite{Sundaralingam2019ICRA}, the efforts here used PointNet++\cite{Qi2017ArXiv}. PointNet++ is a network architecture that directly consumes point cloud data, learns spatial encodings of each point, and hierarchically captures local structure. Although conceived for vision, this architecture is well suited towards BioTac electrode values and FE displacements, as these constitute point sets that are sparse, highly nonuniform, or both.

First, the network architecture was used on purely experimental data. Datasets were split, with approximately 80\% of the indentation trajectories apportioned to training and 20\% to testing. To properly evaluate generalizability, data from each trajectory was kept contiguous; thus, the testing set exclusively contained data from unseen indentations. (The non-contiguous case is presented as well in the Results section.) Target data was normalized to $[0, 1]$ before training. 

Regressions were independently performed \textit{from} electrode coordinates ($19 \times 3$) and values ($19 \times 1$), used as features for each of the points,  \textit{to} contact location ($3 \times 1$) and force vector ($3 \times 1$) using TensorFlow. Training was conducted on an NVIDIA GPU Cloud (NGC) instance and executed for $24$-$36$ hours per dataset. The network consisted of three set abstraction layers followed by two fully connected layers with $1024$ channels. The abstraction layers downsampled the number of points to $32$, $8$, and $1$ by considering points within a radius of $2.5$ mm, $7$ mm, and $\infty$, respectively. To extract feature representations, fully connected layers were applied to the features of each point and others within its radius. Each set abstraction layer used three fully connected layers with dimensions of $[64,64,128]$, $[128,128,256]$, and $[256,256,512]$, respectively.

\subsection{Tactile Field Estimation}

Procedures for estimating tactile fields were nearly identical to those for tactile features. Regressions were now independently performed \textit{from} electrode location coordinates and activation values \textit{to} FE nodal displacements ($N \times 3$) on the BioTac skin mesh, where $N$ is the number of selected nodes. The skin mesh contained over $4000$ total nodes in each simulation; for efficiency, $128$ nodes were selected by randomly sampling along the ventral surface of the skin.

\section{Results}

\subsection{Finite Element Model Validation}

\begin{figure}[thpb]
  \centering
  \framebox{\parbox{3.25in}{\includegraphics[scale=0.97]{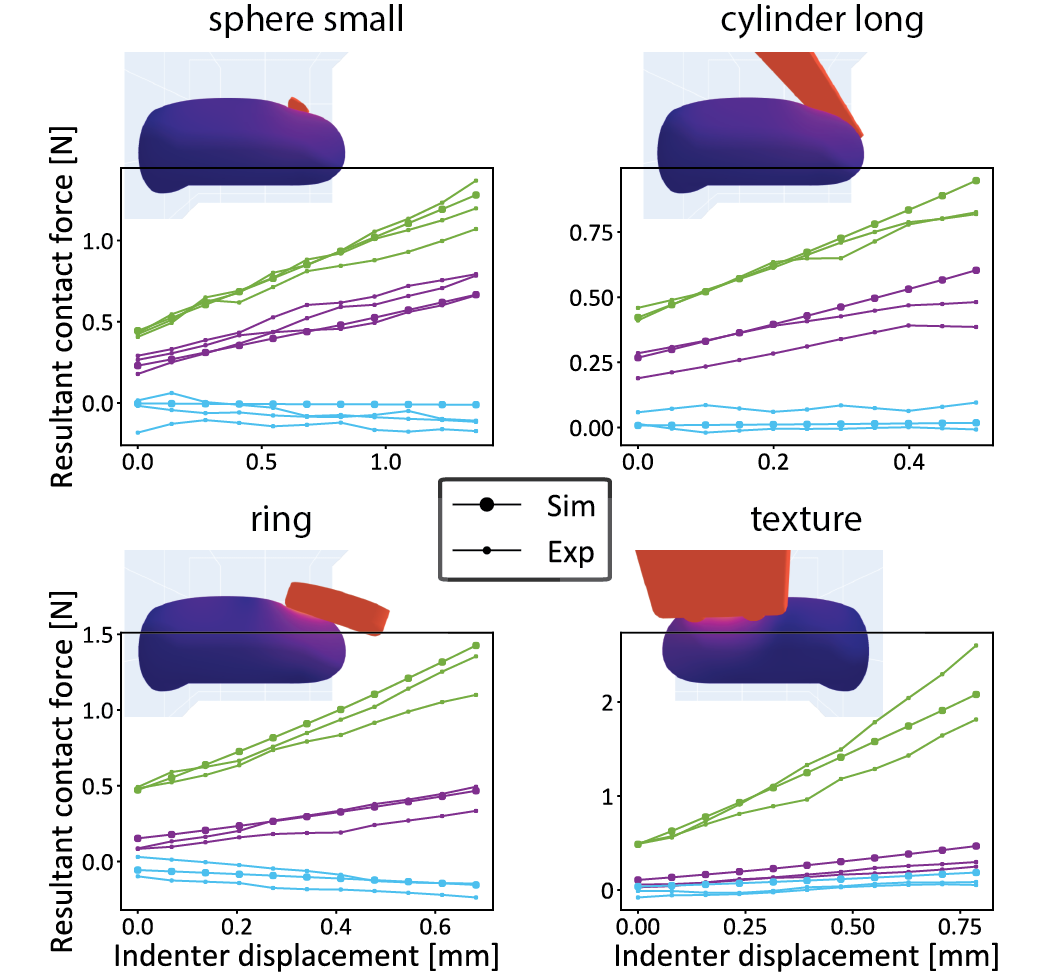}}}
  \caption{Examples from FE model validation. Resultant contact forces for simulation and experiments are shown along the three spatial axes for four high-deformation indentation trajectories, each from a different indenter. Experimental data for three different BioTac sensors is shown. Note that y-axis ranges are different across plots; force-deformation derivatives (i.e., stiffnesses) vary strongly with indenter type and contact location.}
  \label{fe_valid}
\end{figure}

As described earlier, the free parameters in the FE model were calibrated by minimizing the error between predicted and experimental net forces for a \textit{single} angled trajectory of the \textit{sphere medium} indenter. Following calibration, simulation predictions were compared to experimental data over \textit{all} nine indenters and $700$+ trajectories. For each trajectory, RMS error was computed using Equation~\ref{eq:1}.  This process was repeated for all trajectories corresponding to a particular indenter, and the RMS errors were then averaged across the trajectories. The average RMS error ranged from a minimum of $0.226$~$N$ for the \textit{texture} indenter to a maximum of $0.486$~$N$ for \textit{ring}, with a median of $0.260$~$N$. However, different indenters tend to impart different force levels; when RMS errors were normalized by force magnitude range (between $1$-$10$~$N$), predictions for \textit{cylinder small} were most accurate, whereas those for \textit{sphere small} were least accurate. Fig. \ref{fe_valid} compares FE predictions and experimental data for four different indenters.

Given possible sources of experimental error (e.g., BioTac manufacturing variability and F/T sensor noise), as well as the minimal calibration procedure (i.e., using data from a single indentation out of $700$+ possible trajectories), the low error across all indenters demonstrates that the FE model is both sufficiently accurate and highly generalizable.

\subsection{Tactile Feature Estimation}

\subsubsection{Contact Location}

\begin{figure}[thpb]
  \centering
  \framebox{\parbox{3in}{\includegraphics[scale=0.85]{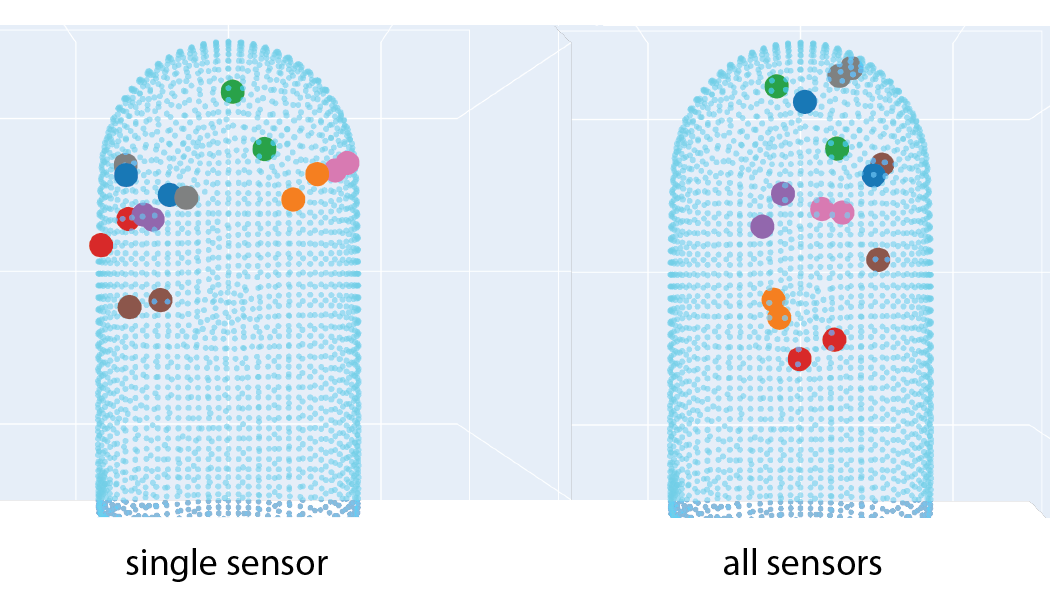}}}
  \caption{Examples from contact location estimation. Left: 8 randomly-sampled contact-location estimates for a network trained on data from a single sensor. Same colors denote target-prediction pairs. Right: 8 additional estimates for a network trained on data from all sensors.}
  \label{contact}
\end{figure}

As previously described, regression from electrode values to tactile features was performed on training/test sets for three individual BioTac sensors, as well as for all the BioTac sensors combined. Mean and median errors in contact location for the BioTac 1\footnote{Recall that BioTac 1 contained data  for three out of the nine indenters (\textit{sphere} shapes) due to subsequent damage. Training and testing on the spheres alone likely contributed to BioTac 1's improved performance.}, 2, 3, and combined test sets were ($1.43$, $1.26$)~$mm$, ($2.65$, $2.24$)~$mm$, ($2.87$, $2.47$)~$mm$, and ($2.45$, $2.08$)~$mm$, respectively (see Fig. \ref{contact} for illustration). These values are similar to the spacing between BioTac electrodes ($1.4$-$2.1$~$mm$) and the threshold for human 2-point discrimination at the fingertip ($2.3$~$mm$)\cite{Won2017JApplOralSci}, suggesting that the estimation framework may be used for object identification and manipulation of small finger-held objects.

Furthermore, it should be noted that the BioTac 2, 3, and combined datasets contained data for large indenters with distinct edges (e.g., \textit{cylinder long, cube}). Contact location was defined as the \textit{center of the tip} of the indenter, but during angled trajectories, these indenters frequently made initial contact at an \textit{edge}. Thus, the worst-case error of $2.87$~$mm$ was substantially better than anticipated, as the regression was occasionally required to infer locations $5$-$10$~$mm$ away. 

\subsubsection{Resultant Force Vector}

\begin{figure}[thpb]
  \centering
  \framebox{\parbox{2.25 in}{\includegraphics[scale=0.8]{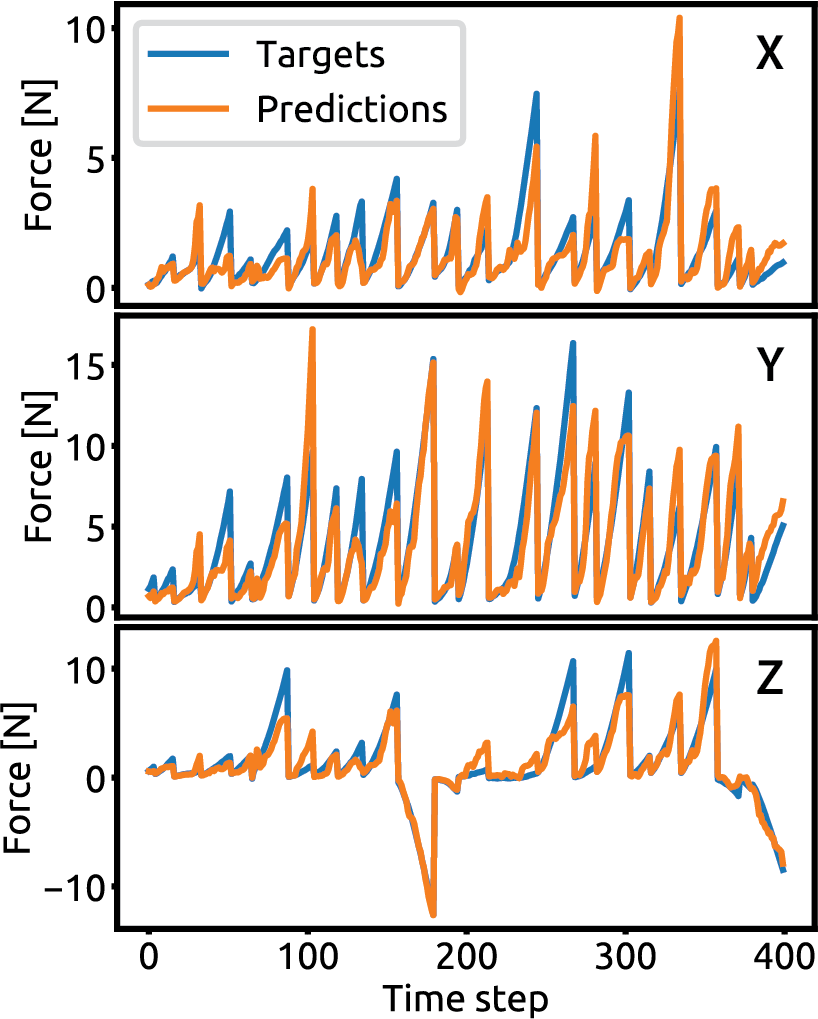}}}
  \caption{Examples from 3D force vector estimation. Force component estimates are shown for a randomly-sampled contiguous slice from the test set. The network here was trained on data from a single sensor. Each spike corresponds to a maximum indentation during a particular trajectory.}
  \label{force}
\end{figure}

For the BioTac 1, 2, 3, and combined test sets, median errors in force magnitude were ($0.62$, $0.54$, $0.50$, $0.56$)~$N$, with mean errors $0.6$-$0.9$~$N$ higher. The higher mean errors are due to the high peak force levels examined throughout data collection ($5$-$20$~$N$). Mean angular cosine errors were ($0.17$~$rad$, $0.13$~$rad$, $0.16$~$rad$, $0.14$~$rad$), with median errors $0.02$-$0.04$~$rad$ lower (Fig. \ref{force}).

In comparison, the best force estimates in the literature on ground-truth data sources are median force errors of $0.32$-$0.51$~$N$ and angular cosine errors of $0.07$-$0.36$~$rad$ over smaller peak forces of $1$-$5$~$N$, in the predominantly normal direction\cite{Sundaralingam2019ICRA}. (A median force error of $0.06$~$N$ was also reported, but most training data for that condition was generated from a semi-analytical model.) These force estimates were used in \cite{Sundaralingam2019ICRA} to lift and place multiple objects, including a mustard bottle and paper cup. The similarity of our results suggests our estimation framework may also be effective for such applications, with potential robustness to more object types and larger forces, as well as the ability to predict shear.

\subsection{Tactile Field Estimation}

\begin{figure}[thpb]
  \centering
  \framebox{\parbox{3.15in}{\includegraphics[scale=0.9]{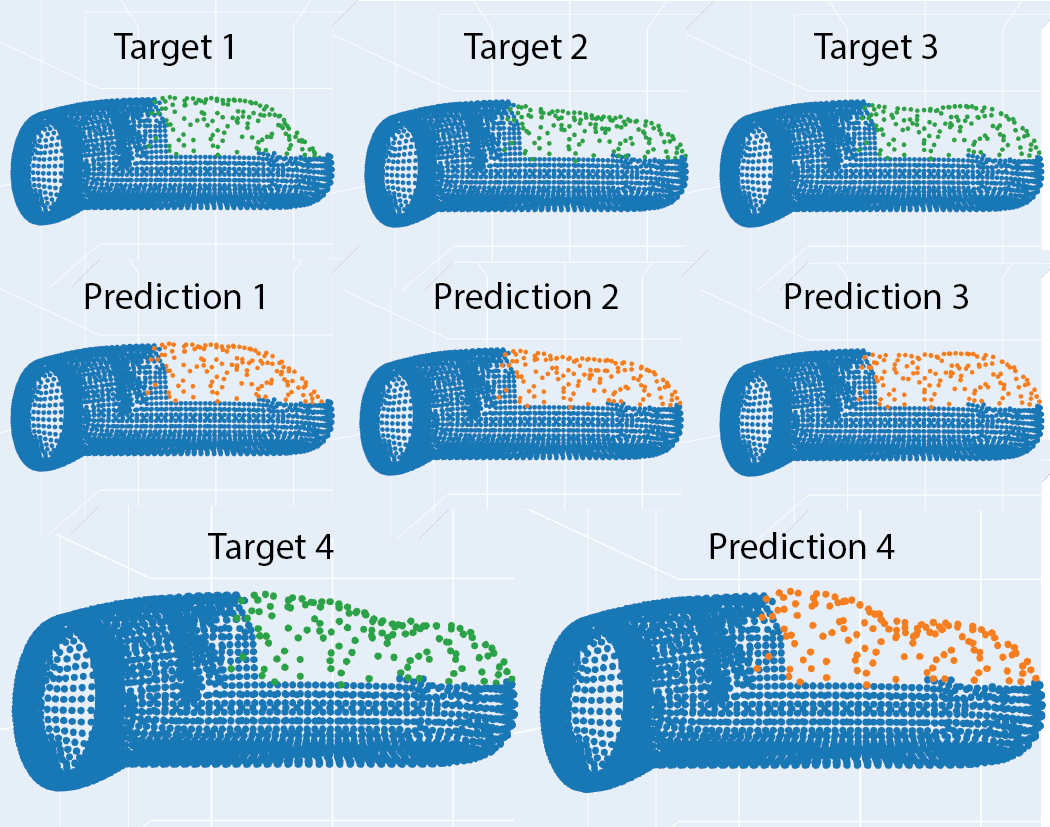}}}
  \caption{Examples of FE nodal displacement estimation. Displacement fields were randomly sampled from the dataset. Most predicted fields were indistinguishable from targets; several other cases with extreme deformations are shown here. Each target-pair prediction is from a different test set.}
  \label{node_disps}
\end{figure}

As with tactile feature estimation, regression from electrode values to tactile fields (here, FE nodal displacements) were performed on the BioTac 1, 2, 3, and combined datasets. The mean nodal displacement errors were ($0.21$, $0.20$, $0.25$, $0.22$)~$mm$, respectively (Fig. \ref{node_disps}). Thus, fingertip deformation fields were well-predicted over the dataset. Visually, complex deformations were captured, and for moderate-to-high indenter displacements (and thus, strong electrode signals), many deformation fields became indistinguishable from FE predictions. 

This result represents a notable departure from prior studies, as estimated information for a contact interaction is not just expressed by a single center-of-pressure value or resultant force vector, but by a high-resolution deformation field. From an abstract perspective, the ability to predict these fields supports the long-discussed hypothesis that the BioTac electrode signals do, in fact, contain this information. Practically speaking, this high-density information enables the perception of geometric features such as flat surfaces, edges, corners, and divots, as visible in several of the deformed surfaces of Fig. \ref{node_disps}. Furthermore, the generation of these fields provides much greater interpretability of the BioTac's cryptic response to contact interactions, facilitating the development and debugging of control algorithms that consume the raw electrode data.

\subsection{Experimental Data Contiguity}

As described previously in the Methods section, experimental data from each trajectory was kept contiguous when apportioning the training and testing datasets, such that the test set contained only data from unseen indentations. As a validation experiment, the non-contiguous case was also considered, where individual data points were randomly sampled when apportioning the datasets. As anticipated, estimation accuracy on the test set dramatically improved: mean contact location error decreased by $10$x (to approx. $0.2$~$mm$), median force magnitude and angular error decreased by $3$x (to approx. $0.15$~$N$ and $0.05$~$rad$, respectively), and mean nodal displacement error decreased by $2.5$x (to approx. $0.08$~$mm$). These deceptively-low errors reiterate the need for careful dataset construction when estimating tactile quantities.

\subsection{Estimation for Unseen Sensors and Indenters}

As described earlier, despite being calibrated using data from a single indentation, the FE model generalized strongly across BioTac sensors, as well as the wide range of indenters and trajectories (as illustrated in Fig.~\ref{fe_valid}). An analogous examination of generalizability can be considered for the neural-network estimation framework for purely experimental data.

First, generalizability to unseen sensors was investigated. Networks trained on experimental data from BioTac 2 were tested on data from BioTac 3, and vice versa. (BioTac 1 was not included due to the decreased number of indenters used in experimental testing.) Errors in contact location, force magnitude, and force angle increased by factors of approximately $2$-$3$ relative to testing on the original BioTac. Thus, the networks generalized weakly from one sensor to another. Nevertheless, the increase in estimation error was fully expected; during extensive experimental testing of four different BioTac sensors, electrode responses were found to be quantitatively and qualitatively distinct across sensors, even under highly-controlled testing conditions. These variations may be a result of substantial manufacturing variability in the BioTac's electronic components, unavoidable to the user.

Next, generalizability to unseen indenters was examined. Leave-one-out analysis was performed for BioTac 2 and 3, in which networks were trained on data from eight of the nine indenters and tested on data from the missing indenter. When the unseen indenter was geometrically similar to others in training (e.g., if the unseen indenter was a \textit{sphere} or \textit{cylinder} type), errors in tactile features were largely unaffected; for example, contact location and force angle were nearly constant at approximately $2$~$mm$ and $0.15$~$rad$, respectively. However, when the indenter was geometrically \textit{dissimilar} to others in training, errors increased. For example, for the geometrically-unique \textit{ring}, mean prediction errors for contact location and force angle increased by $1.5$~$mm$ and $0.05$ ~$rad$. Given the dependence on geometric similarity, firmly generalizing the prediction of tactile quantities to arbitrary object geometries may require training on an even greater number of indenters.

In order to effectively leverage the experimental dataset for direct estimation of tactile quantities, it is highly recommended to use the data to pretrain a neural network, after which fine-tuning can be performed for a particular sensor. The same procedure can also be applied for unseen objects; however, due to the deliberate design of the indenters as shape primitives of everyday items (Fig. \ref{indenters}), such steps may not prove necessary.

\section{Conclusions}

This research advances the modeling and interpretation of tactile sensors, with a focus on the SynTouch BioTac sensor due to its high performance and widespread use in research. The first contribution was a precise, diverse experimental dataset for the BioTac, consisting of three BioTac sensors, nine indenters, and over 300 unique indentation trajectories, $700$ total trajectories, and $30k$ data points after processing; additionally, approximately $70\%$ of trajectories were designed to induce shear. Next, the study provided the first ever 3D FE model of the BioTac. The model implemented a hyperelastic constitutive law for the skin, captured fluidic incompressibility, and simulated frictional contact across multiple surfaces. The accuracy and  generalizability of the model was validated across the wide range of physical experiments. Finally, this research closely matches previous benchmarks in estimating contact location and 3D force vector, and more importantly, extends the boundaries of tactile estimation by presenting the first accurate predictions of tactile \textit{fields} from raw BioTac data.

These results are important for the advancement of tactile sensing in robotics research for multiple reasons. First, the dataset captures diverse contact interactions beyond existing efforts. For example, previous studies typically focused on manually contacting the BioTac with one to two indenters predominantly (or exclusively) in directions normal to the sensor surface. Additionally, the relative pose of the indenter and BioTac was rarely measured with precision, if at all. Nevertheless, most real-world objects have wide geometric diversity; shear is highly critical for slip detection, mass detection, palpation, and sliding or spreading objects; and knowledge of contact location is invaluable for grasping and dexterous manipulation. Second, as recently initiated for vision-based tactile sensors\cite{Ma2018ICRA}, an accurate FE model of the BioTac enables previously-inaccessible capabilities. Researchers can directly use the model to predict deformation fields of the sensor and object, as well as contact force distributions transmitted through the sensor-object interface. These predictions can be leveraged to provide more accurate assessments of stability with soft contact, prevent damage to brittle or delicate materials (like fruits or living tissue), and guide the purposeful reshaping of materials (e.g., the flattening of dough). Finally, the ability to regress to tactile fields directly from electrode values enables the rich information in the FE model to be accessible from raw signals at runtime.

Future work will focus on key applications and remaining challenges. For instance, the BioTac sensors will be attached to dexterous robotic manipulators, and the learned regressions from electrode values to contact forces and displacement fields will be utilized for surface- and contour-following, which constitute essential exploratory actions for numerous tasks. In addition, expanding on an initial effort\cite{Ruppel2019IAS}, these regressions will be inverted in order to predict raw electrode signals from FE deformation fields. By integrating inverse models into robotic simulators such as MuJoCo, Gazebo, Drake, or NVIDIA Isaac, realistic electrode responses can be generated in simulated environments, facilitating the training of control policies that take full advantage of rich contact information.

\section*{Acknowledgments}

The authors would like to thank Krishna Mellachervu of ANSYS Inc. for outstanding technical support; Keunhong Park for numerous insightful discussions; Jeremy Fishel of SynTouch Inc. for providing BioTac CAD models and preliminary mechanical data on the rubber skin; and Balakumar Sundaralingam for providing data from his previous analyses.

\bibliographystyle{plainnat}
\bibliography{refs}

\end{document}